\documentclass[letterpaper]{article} 
\usepackage{aaai2027}  
\usepackage[hyphens]{url}  
\usepackage{graphicx} 
\usepackage{natbib}  
\usepackage{caption} 
\usepackage{algorithm}
\usepackage{algorithmic}
\usepackage{newfloat}
\usepackage{listings}
\DeclareCaptionStyle{ruled}{labelfont=normalfont,labelsep=colon,strut=off} 
\floatstyle{ruled}
\newfloat{listing}{tb}{lst}{}
\floatname{listing}{Listing}

\usepackage{booktabs}
\usepackage{amsmath}
\usepackage{amssymb}

\ifdefined
\fi

\title{Operation-aware Multimodal Clinical Decision Support System \\ for ICU Discharge Risk Assessment}

\author{
Guoqing Zhang,\textsuperscript{\rm 1}
Rafik Hadfi,\textsuperscript{\rm 1}
Takayuki Ito\textsuperscript{\rm 1}
}
\affiliations{
\textsuperscript{\rm 1}Graduate School of Informatics, Kyoto University, Kyoto, Japan \\
zhang.guoqing.27t@st.kyoto-u.ac.jp, rafik.hadfi@i.kyoto-u.ac.jp, ito@i.kyoto-u.ac.jp
}

\usepackage{multirow}
\usepackage{tabularx}
\usepackage{array}
\usepackage{xcolor}
\usepackage{fontawesome5}
\usepackage{tikz}
\usetikzlibrary{shapes.geometric, arrows.meta, positioning, calc, backgrounds, fit}

\definecolor{promptbg}{HTML}{F8F9FA}
\definecolor{promptborder}{HTML}{D0D7DE}
\definecolor{cIndigo}{HTML}{283593}
\definecolor{cTeal}{HTML}{00695C}
\definecolor{cPurple}{HTML}{6A1B9A}

\begin{document}

\maketitle

\begin{abstract}
    Unplanned 30-day intensive care unit (ICU) readmission following hospital discharge is a major benchmark for evaluating healthcare quality and patient safety in modern critical care medicine. Most existing predictive approaches assume that readmission risk is determined solely by a patient's individual pathophysiological state, ignoring the institution's resource capacity and operational context (Hospital Operational Context). In clinical practice, discharge decisions are frequently dynamic trade-offs made by intensivists balancing patient medical stability against ward bed turnover pressures. To address this gap, we design an Operation-aware Multimodal Clinical Decision Support System (CDSS) for ICU discharge risk assessment. On the one hand, the system unites multi-agent collaborative extraction with self-evolving retrieval-augmented generation (RAG) to eliminate formatting-template noise across lengthy progress notes. On the other hand, it constructs departmental workload indicators (\texttt{icu\_speedup\_los}) alongside admission flow characteristics, unifying continuous vital signs, hierarchical sparse medical codes (ICD/DRG), and discharge summary semantics. Addressing the extreme class imbalance ($\sim$10.5\% readmission incidence) and the challenge of alarm fatigue, the system implements data normalization, class reweighting optimization, and an F1 decision-threshold calibration pipeline. In 20 independent multi-seed runs conducted on a cohort of 20,000 ICU stays from MIMIC-III, empirical findings show that: a single-modality baseline depending exclusively on vital signs possesses limited predictive capacity under real-world class imbalance (PR-AUC of only $0.1387 \pm 0.0086$); directly concatenating raw text embeddings causes performance degradation in decision tree models (XGBoost and LightGBM PR-AUC drop to 0.3893 and 0.4030, respectively); whereas an end-to-end multimodal fine-tuning architecture (Transformer E2E FineTune) utilizing layer-wise parameter freezing and differential learning rates successfully reverses text negative transfer, achieving the overall highest AUROC of $0.8288 \pm 0.0110$, PR-AUC of $0.4373 \pm 0.0310$, and a competitive F1-Score of $0.4757 \pm 0.0196$. Mechanistic attribution analyses also confirm that operational indicators supply independent risk signals, establishing an engineering foundation for deploying high-utility clinical decision support tools in bedside hospital wards.
    
    \noindent\textbf{Code} --- \url{https://github.com/KuoChing-cmd/ClinicalNoteAgent/}
\end{abstract}

\section{Introduction}

In intensive care clinical decision-making, identifying vulnerable patients at risk of latent deterioration and avoiding premature step-down to general wards are core imperatives of critical care medicine. Patients experiencing unplanned ICU readmission within 30 days of discharge undergo substantially prolonged hospital stays and face elevated in-hospital mortality \citep{klinge2020readmission, barbieri2020benchmarking, sheetrit2023predicting}. While recent multicenter investigations confirm the promise of machine learning for post-discharge surveillance \citep{lim2025multicenter}, existing clinical early warning systems commonly encounter barriers during ward deployment due to alarm fatigue: in low-prevalence locales where readmissions occur in only $\sim$10\% of cases, typical models optimized for high AUROC tend to trigger substantial false-positive alarms, prompting clinicians to turn off the alert systems. To construct an actionable Clinical Decision Support System (CDSS), one must preserve screening sensitivity while substantially improving precision, thereby driving PR-AUC and F1 scores toward clinically acceptable operating points.

Although electronic health record (EHR) repositories such as MIMIC-III \citep{johnson2016mimic} have advanced machine learning in critical care---spanning deep temporal focus \citep{madinei2024attention}, graphical EHR representations \citep{choi2020learning}, and recent readmission benchmarks \citep{fathy2025comprehensive}---state-of-the-art predictive models still encounter two fundamental deployment bottlenecks.
First, they exhibit a neglect of hospital operational context. Standard models conceptualize patients as closed physiological entities, presuming that deterioration following transfer stems entirely from built-in disease progression. In clinical practice, however, ICU discharge is fundamentally a resource deployment and scheduling process carried out under institutional bed constraints \citep{kim2015icu, wagner2013outcomes, long2018boarding}. Intensivists must routinely weigh incomplete physiological recovery against acute bed demands driven by emergency department (ED) boarding or post-operative transfers. Patient transfer frequency before ICU admission, ED boarding durations, and ward acceleration driven by bed saturation directly mirror operational workflow strain and care coordination friction; yet, customary doctrines remain entirely agnostic to this systemic background.
Second, models suffer from data disparity and text template noise. Intensive care records span high-frequency temporal streams, sparse diagnostic codes, and lengthy unstructured progress notes. While recent studies explore medical text via knowledge graphs \citep{lu2021predicting} or multimodal architectures for readmission prediction \citep{helmy2025leveraging, datta2025improving}, clinical free-text is often lengthy, laden with jargon, and burdened by copy-pasted administrative templates. Directly feeding unadapted high-dimensional text embeddings fails to yield meaningful gains and instead introduces severe feature interference (negative transfer) with strong baselines such as gradient-boosted decision trees.

To resolve these problems, we reframe ICU discharge evaluation as an operationally aware multimodal decision-support task and develop a CDSS architecture that encompasses self-evolving textual distillation and deep cross-modal correlation. The core contributions of this paper are summarized as follows:
\begin{itemize}
    \item Operation-aware Decision Framework: Overcoming randomized timestamp offsets in MIMIC-III, we introduce unit-benchmarked turnover deviation indicators (\texttt{icu\_speedup\_los}) and care flow complexity features, eliminating the subjective assumption that outcomes depend solely on physiological status.
    \item Self-Evolving Multi-Agent Text Distillation: We formulate a 7-domain specialist agent collaborative division of labor paired with a reflective evolutionary experience repository, accumulating 3,085 rules across 5,000 patient samples. Combined with FAISS vector retrieval for structured information distillation, we propose a layer-wise, frozen, end-to-end fine-tuning scheme that systematically mitigates interference from template noise in clinical notes.
    \item Thorough Testing under Low-Prevalence Conditions: Across a cohort of 20,000 ICU stays, we conduct 20 independent randomized multi-seed runs (Seeds 42--61), exposing the weakness of single-modality physiological prediction (PR-AUC of only 0.1387) and verifying the robustness of the end-to-end fine-tuning architecture in achieving global optimality (AUROC 0.8288, PR-AUC 0.4373, F1 0.4757).
    \item Interpretability \& Clinical Utility Analysis: Through integrated gradients \citep{sundararajan2017axiomatic}, we systematically explain the mechanistic contributions of separate physiological channels, pivotal time-based attention windows, and operational flow variables in 30-day readmission forecasting.
\end{itemize}

\section{Clinical Workflow and Problem Formulation}

\subsection{Clinical Prediction Task and Endpoint Definition}
The prediction target is unplanned ICU readmission within 30 days following discharge (30-Day Readmission). In light of complex intra-hospital patient trajectories, endpoint labels $y_i \in \{0, 1\}$ follow strict clinical mutually exclusive criteria:
\begin{itemize}
    \item Cross-admission readmission: The patient undergoes a separate hospital admission and enters an ICU within 30 days post-discharge (same \texttt{SUBJECT\_ID}, distinct \texttt{HADM\_ID}).
    \item Within-admission readmission: During the same hospitalization episode, the patient is transferred out of the ICU, deteriorates, and is subsequently readmitted back to the ICU (multiple ICU stays under the same \texttt{HADM\_ID}, where the subsequent admission timestamp postdates the preceding transfer timestamp, i.e., $\text{INTIME}_{k+1} > \text{OUTTIME}_k$).
    \item Mortality Mutual-Exclusion: Patients who suffer in-hospital mortality (\texttt{HOSPITAL\_EXPIRE\_FLAG = 1}) have their readmission risk strictly calibrated to 0, eliminating logical contradictions.
\end{itemize}

\subsection{Multimodal Clinical Data Streams and Domain Alignment}
For each ICU stay event $i \in \{1, \dots, N\}$ (experimental cohort $N = 20,000$), the input is structured as a heterogeneous quadruple:
\begin{equation}
    \mathcal{X}_i = \left\{ \mathbf{X}_i^{seq}, \mathbf{x}_i^{stat}, \mathbf{x}_i^{code}, \mathbf{x}_i^{text} \right\}
\end{equation}

First, for high-frequency dynamic physiological sequences ($\mathbf{X}_i^{seq} \in \mathbb{R}^{48 \times 8}$), the system captures 8 core physiological signs during the initial 48 hours following ICU admission: Heart Rate (HR), Respiratory Rate (RR), Oxygen Saturation ($\text{SpO}_2$), Systolic Blood Pressure (SBP), Diastolic Blood Pressure (DBP), Mean Arterial Pressure (MAP), Temperature (Temp), and Blood Glucose (Glucose). Blood pressure measurements explicitly distinguish invasive arterial pressure from non-invasive cuff pressure; when both coexist, the invasive arterial line---the gold standard in critical care---is preferred. Time series are resampled into 1-hour regularized bins, with missing steps forward-filled.

Second, structured demographic and operational indicators ($\mathbf{x}_i^{stat} \in \mathbb{R}^{15}$) capture baseline patient context and care delivery friction across three facets:
(i)~\textit{Baseline \& Surgical covariates} include age, gender, marital status, ethnicity, insurance, admission source/type, first ICU unit, surgical procedure count, surgery binary indicator, and log-transformed surgery-to-ICU entry gap $\ln(1 + \Delta t_{\text{surg}})$.
(ii)~\textit{Transfer Patterns} track the number of ward transfers prior to ICU admission (\texttt{pre\_icu\_transfers}) along with log-transformed emergency department wait time $\ln(1 + t_{\text{ED}})$.
(iii)~\textit{Ward Pressure \& Turnover Deviation}: addressing the obstacle where randomized absolute timestamps in MIMIC-III preclude concurrent bed occupancy observation across patients, we formulate a stay deviation metric aligned with historical unit benchmarks (e.g., MICU, SICU) along with stay duration features:
\begin{equation}
    \begin{aligned}
        \text{icu\_speedup\_los} & = \overline{\text{LOS}}_{\text{unit}} - \text{LOS}_{\text{actual}} \\
        \text{log\_icu\_los}     & = \ln(1 + \text{LOS}_{\text{hours}})
    \end{aligned}
\end{equation}
Specifically, under a zero-leakage cutoff, the feature is strictly computed at the exact moment of the step-down transfer order, sharing the same temporal cross-section with discharge notes and precluding future information leakage. Crucially, contrary to the naive assumption that shorter stays merely reflect milder acuity (which would yield lower readmission rates), empirical attribution reveals that positive \texttt{icu\_speedup\_los} (early turnover) significantly correlates with higher readmission risk (20.1\% continuous IG attribution). This refutes indication confounding and confirms that the metric captures external operational bed-turnover pressure and premature discharge before complete physiological stabilization.

Third, hierarchically decoupled sparse medical codes ($\mathbf{x}_i^{code}$) capture hierarchical EHR coding structures \citep{choi2020learning}. Primary ICD diagnoses and primary DRG codes select the Top-128 categories, mapped into categorical indices feeding a learnable continuous embedding layer (128 dimensions). Secondary diagnoses (Top-32), surgical procedures (Top-64), and prescribed medications (Top-64) are encoded into multi-hot sparse vectors, projected through embedding weight matrices, mean-pooled across active entries, and concatenated into compact representations.

Fourth, unformatted clinical progress representations ($\mathbf{x}_i^{text}$) incorporate free-text clinical progress notes. In contrast to standard clinical text baselines \citep{lu2021predicting, helmy2025leveraging} that directly process progress notes vulnerable to template noise, lengthy inpatient records are distilled via self-evolving multi-agent workflows into structured summaries, with 768-dimensional semantic features extracted by \texttt{Bio\_ClinicalBERT} \citep{alsentzer2019publicly}.

\section{Operation-aware CDSS Architecture}

\begin{figure*}[t]
    \centering
    \includegraphics[width=0.78\textwidth]{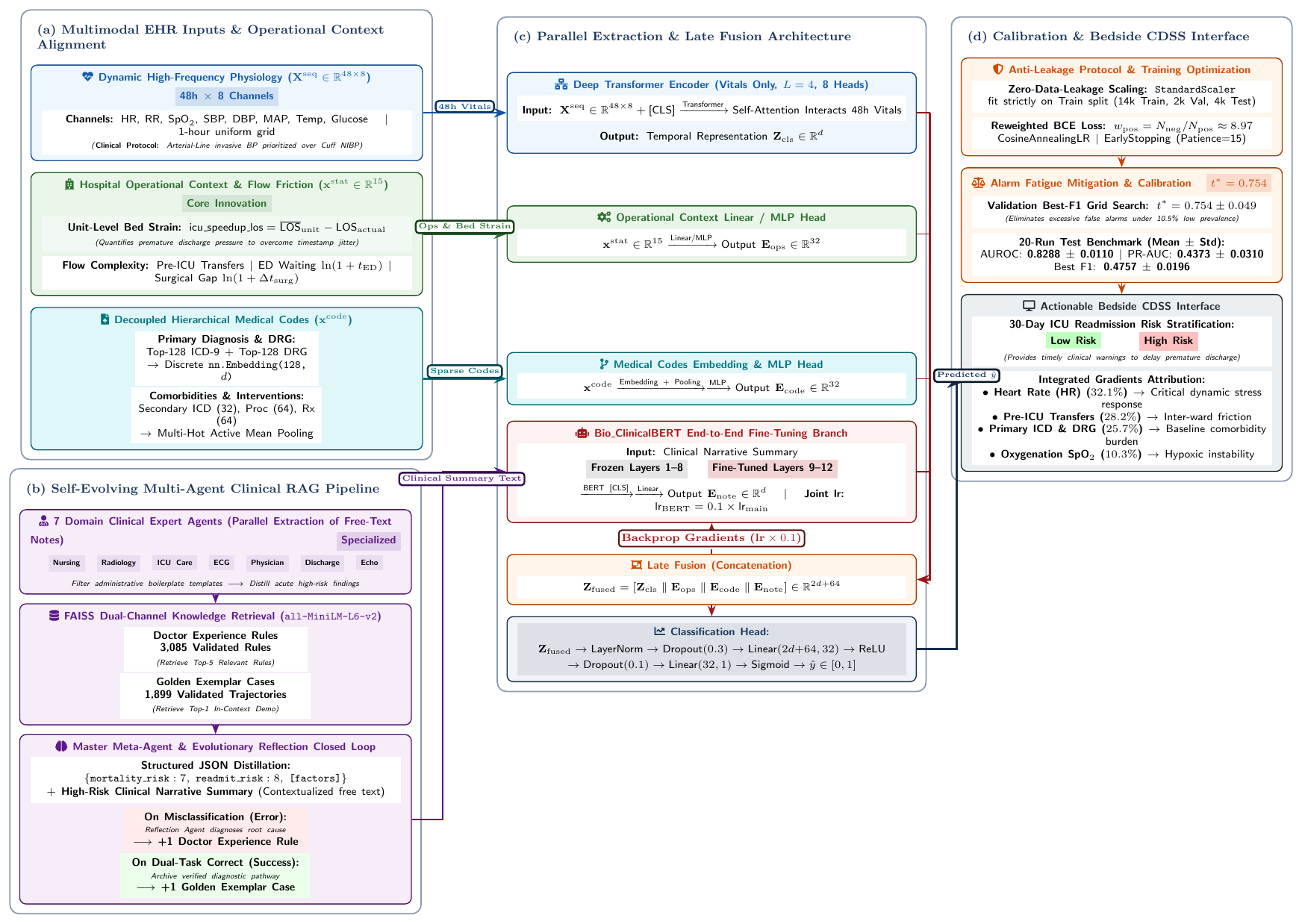}
    \caption{System overview of the Operation-aware Multimodal Clinical Decision Support System (CDSS). (a) Multimodal EHR inputs aligned with hospital operational bottleneck perception; (b) Self-evolving multi-agent clinical RAG mining pipeline with 7 domain specialists and dual-track rule/case repositories; (c) Parallel feature derivation and Late Fusion architecture alongside the \texttt{Bio\_ClinicalBERT} end-to-end fine-tuning pathway; (d) Zero-leakage normalization, alarm-fatigue calibrated thresholds, and bedside risk-stratified decision support.}
    \label{fig:framework}
\end{figure*}

\subsection{Unstructured Text Mining: Multi-Agent Self-Evolving RAG Mechanism}
Orchestrated multi-agent architectures sustain diagnostic fidelity under complex clinical workloads compared to monolithic models \citep{klang2026orchestrated}. To overcome verbosity and repetitive administrative boilerplate in clinical progress notes, the system establishes a self-evolving multi-agent distillation pipeline across five coordinated stages:
(i)~\textit{Hierarchical Specialist Collaboration}: Clinical text is segmented across 7 specialized clinical domains (Nursing, Radiology, ICU Care, ECG, Attending Physician, Discharge Summary, and Echocardiography), where independent domain agents extract acute warning signals.
(ii)~\textit{Meta-Agent Synthesis}: A supervisory Meta-Agent aggregates domain summaries, synthesizing a structured JSON report containing a prioritized vital factor roster and composite risk score.
(iii)~\textit{Reflective Evolution Loop}: When predictions diverge from clinical ground truth during training, a Reflection Agent audits causal failure points, extracting physician-vetted heuristic rules into a knowledge base (3,085 rules); correctly classified trajectories are cataloged as gold-standard exemplars (1,899 cases).
(iv)~\textit{Lightweight RAG Retrieval}: Guided by \texttt{all-MiniLM-L6-v2} embeddings and a FAISS index, the pipeline dynamically retrieves the Top-5 relevant heuristic rules and Top-1 exemplar case per admission as few-shot in-context demonstrations.
(v)~\textit{Feature Embedding \& Projection}: Distilled summaries are encoded by \texttt{Bio\_ClinicalBERT} into 768-dimensional contextual vectors and projected to a $d$-dimensional embedding $\mathbf{E}_{\mathrm{note}}$ aligned with temporal representations.

\subsection{Cross-Modal Network Architecture and Deep Fusion Schemes}
The framework investigates three architectural paradigms.
First, we evaluate tree-based baselines: XGBoost \citep{chen2016xgboost} and LightGBM \citep{ke2017lightgbm} fitted on physically concatenated vectors combining temporal statistical aggregations, static covariates, multi-hot codes, and 768-dimensional text embeddings \citep{chen2025xgboost}.
Second, we construct sequential late fusion (LSTM Late Fusion): a 4-layer bi-directional LSTM with self-attentive temporal pooling encodes physiological sequences, concatenated at the classification tier with static, code, and text representations projected via independent MLP branches.
Third, we formulate parallel multi-branch extraction and late fusion. Dynamic physiological sequences are encoded via a deep Transformer encoder \citep{vaswani2017attention} into $\mathbf{Z}_{\mathrm{cls}}$; static operational indicators and medical codes are mapped via independent MLPs into dense vectors $\mathbf{E}_{\mathrm{ops}}$ and $\mathbf{E}_{\mathrm{code}}$. These representations are concatenated with text embedding $\mathbf{E}_{\mathrm{note}}$ to form $\mathbf{Z}_{\mathrm{fused}} = [\mathbf{Z}_{\mathrm{cls}} \parallel \mathbf{E}_{\mathrm{ops}} \parallel \mathbf{E}_{\mathrm{code}} \parallel \mathbf{E}_{\mathrm{note}}] \in \mathbb{R}^{2d+64}$, feeding a 2-layer MLP head for joint risk estimation. To prevent representational collapse during end-to-end training, the initial 8 layers of \texttt{Bio\_ClinicalBERT} are frozen, leaving deeper layers (Layers 9--12) active; BERT parameters receive a 0.1$\times$ scaled learning rate (\texttt{lr * 0.1}) with gradient checkpointing enabled.

\subsection{Training Optimization and Leak-Free Validation Protocols}
To ensure leak-free evaluation under severe class imbalance ($\sim$10.5\% prevalence) and achieve stable multimodal convergence, we institute four disciplined methodological safeguards:
(i)~\textit{Zero-Leakage Normalization}: \texttt{StandardScaler} parameters are fitted strictly on training splits; validation and testing cohorts inherit these statistics without re-estimation, precluding distribution leakage.
(ii)~\textit{Dynamic Cost-Sensitive Loss}: To counter minority-class sparsity, a weighted binary cross-entropy loss deploys dynamic inverse class ratio weighting derived strictly from the training split: $w_{\mathrm{pos}} = N_{\mathrm{neg}} / N_{\mathrm{pos}} \approx 8.97$.
(iii)~\textit{Cosine Scheduling \& Rollback}: A \texttt{CosineAnnealingLR} schedule regulates learning rate decay, paired with early stopping (\texttt{patience=15}) on validation loss; upon termination, model parameters rollback to the checkpoint achieving the minimal validation loss.
(iv)~\textit{Decision Threshold Calibration}: On validation splits, exhaustive grid search (step size 0.01) determines the optimal decision cutoff $t^*$ maximizing the $F_1$-Score, replacing the uncalibrated default 0.5 threshold for final test evaluation.

\begin{table*}[t]
    \centering
    \small
    \caption{30-day unplanned ICU readmission prediction performance on the test set (20 independent runs, mean $\pm$ Std, evaluated @ optimal F1 decision threshold). Bold indicates global optimum.}
    \label{tab:main_results}
    \resizebox{\textwidth}{!}{
        \begin{tabular}{llcccc}
            \toprule
            Model Category & Model Name \& Variant                                                   & Optimal Threshold $t^*$ & AUROC                        & PR-AUC                       & F1-Score                     \\
            \midrule
            \textit{Tree Baselines \& Text Ablation}
                           & XGBoost Base                                                            & $0.314 \pm 0.093$       & $0.8055 \pm 0.0098$          & $0.4125 \pm 0.0209$          & $0.4465 \pm 0.0193$          \\
                           & XGBoost + LLM Notes                                                     & $0.134 \pm 0.053$       & $0.7914 \pm 0.0118$          & $0.3893 \pm 0.0213$          & $0.4400 \pm 0.0166$          \\
                           & LightGBM Base                                                           & $0.463 \pm 0.067$       & $0.8158 \pm 0.0111$          & $0.4114 \pm 0.0187$          & $0.4555 \pm 0.0172$          \\
                           & LightGBM + LLM Notes                                                    & $0.370 \pm 0.064$       & $0.8087 \pm 0.0114$          & $0.4030 \pm 0.0193$          & $0.4548 \pm 0.0170$          \\
            \midrule
            \textit{Deep Multimodal Networks \& Fusion}
                           & LSTM Base (Pure Structured)                                             & $0.747 \pm 0.051$       & $0.8251 \pm 0.0117$          & $0.4360 \pm 0.0305$          & $\mathbf{0.4798 \pm 0.0200}$ \\
                           & LSTM LateFusion (+ Notes)                                               & $0.726 \pm 0.034$       & $0.8258 \pm 0.0091$          & $0.4338 \pm 0.0301$          & $0.4791 \pm 0.0193$          \\
                           & Transformer EarlyFusion (+ Notes)                                       & $0.757 \pm 0.068$       & $0.8177 \pm 0.0145$          & $0.4278 \pm 0.0332$          & $0.4769 \pm 0.0200$          \\
                           & Transformer LateFusion (+ Notes)                                        & $0.762 \pm 0.042$       & $0.8277 \pm 0.0104$          & $0.4357 \pm 0.0320$          & $0.4751 \pm 0.0158$          \\
                           & LSTM E2E FineTune (+ Notes)                                             & $0.733 \pm 0.044$       & $0.8260 \pm 0.0119$          & $0.4338 \pm 0.0326$          & $0.4754 \pm 0.0178$          \\
                           & Transformer E2E FineTune (+ Notes)                                      & $0.754 \pm 0.049$       & $\mathbf{0.8288 \pm 0.0110}$ & $\mathbf{0.4373 \pm 0.0310}$ & $0.4757 \pm 0.0196$          \\
            \midrule
            \textit{Modality \& Feature Ablation (relative to LSTM Base)}
                           & w/o Admission Dynamics (\texttt{pre\_icu\_transfers}, $t_{\text{ED}}$)  & $0.750 \pm 0.043$       & $0.8260 \pm 0.0103$          & $0.4292 \pm 0.0287$          & $0.4772 \pm 0.0204$          \\
                           & w/o ICU Operational Features (\texttt{icu\_speedup\_los}, $\text{LOS}$) & $0.734 \pm 0.044$       & $0.8275 \pm 0.0111$          & $0.4278 \pm 0.0312$          & $0.4791 \pm 0.0195$          \\
                           & w/o Medical Codes ($\mathbf{x}^{code}$)                                 & $0.741 \pm 0.033$       & $0.8243 \pm 0.0107$          & $0.4334 \pm 0.0302$          & $0.4757 \pm 0.0187$          \\
                           & Only Vitals ($\mathbf{X}^{seq}$)                                        & $0.521 \pm 0.033$       & $0.6232 \pm 0.0149$          & $0.1387 \pm 0.0086$          & $0.2213 \pm 0.0099$          \\
                           & Only Static \& Codes ($\mathbf{x}^{stat}, \mathbf{x}^{code}$)           & $0.735 \pm 0.040$       & $0.8227 \pm 0.0097$          & $0.4294 \pm 0.0273$          & $0.4749 \pm 0.0180$          \\
            \bottomrule
        \end{tabular}
    }
\end{table*}

\section{Experimental Evaluation and Validation}

\subsection{Evaluation Protocol and Cohort Setup}
Experiments are run across 20,000 ICU stay records from the MIMIC-III database. The cohort is partitioned into training (14,000 stays), validation (2,000 stays), and test splits (4,000 stays) in a 7:1:2 ratio, with positive readmission prevalence ranging from 10.5\% to 11.4\%. To eliminate partitioning anomalies, all candidate approaches are evaluated across 20 independent randomized runs (Seeds 42--61), with test performance reported as $\text{Mean} \pm \text{Std}$.

\subsection{Comprehensive Performance Comparison and Paradigm Defense}
Table~\ref{tab:main_results} summarizes the evaluation measures across 15 model configurations evaluated at their optimal F1 decision thresholds across 20 randomized runs. Three core empirical insights emerge.
First, we observe text negative transfer in shallow concatenation. Directly appending unadapted 768-d text embeddings into decision trees (XGBoost/LightGBM) triggers notable performance decline. Inpatient notes are saturated with administrative templates and boilerplate shorthand; high-dimensional embeddings diffuse greedy tree node splitting, diluting compact, high-SNR physiological and operational signals.
Second, results demonstrate the strength of structured baselines and multimodal reframing. Purely structured \texttt{LSTM Base} establishes a solid benchmark ($F_1 = 0.4798$), demonstrating the fitting efficiency of recurrent cells over 48-hour dynamics along with discrete codes. The multimodal goal is therefore not to surpass structured models trivially, but to resolve a harder clinical challenge: \textit{safely incorporating high-value text semantics without suffering negative transfer} (as observed in Transformer EarlyFusion, which degrades under raw text noise).
Third, we achieve a reversal of negative transfer via end-to-end fine-tuning. With layer-wise parameter freezing (Layers 9--12 active) and differential learning rates, \texttt{Transformer E2E FineTune} is the sole architecture reversing text negative transfer into positive gains, securing global peaks in both AUROC ($0.8288$) and PR-AUC ($0.4373$). Paired hypothesis tests across 20 matched random seeds confirm that the AUROC gain of end-to-end fine-tuning over static early fusion is highly statistically significant (mean AUROC 0.8288 vs.\ 0.8177; Paired $t$-test $t = 4.89, p < 0.001$; Wilcoxon $p < 0.001$). This provides strong empirical evidence that end-to-end backpropagation prevents the text representation from being overshadowed by dense temporal dynamics, enabling the language model to better align textual risk cues with structured physiological trajectories.

\subsection{Sensor Dropout Robustness}
In operational ICU environments, electrode detachment, patient movement, and intermittent contact failures occur frequently. To assess fault tolerance under incomplete physiological telemetry, we evaluated the representative \texttt{LSTM LateFusion} model under random telemetry loss (Sensor Dropout from 10\% to 30\%) across 20 independent runs. The system presented remarkable stability: under 30\% channel dropout, AUROC decayed by less than 0.0005 in absolute terms (from $0.8258 \pm 0.0091$ to $0.8254 \pm 0.0092$), PR-AUC remained steady ($0.4338 \to 0.4330$), and $F_1$ was preserved at $0.4783 \pm 0.0189$. This confirms that multimodal fusion provides robust cross-modal safety redundancy, in which static operational indicators, diagnostic codes, and clinical text insulate decision support due to localized telemetry disruptions.

\subsection{Modal Ablation and Vitals Collapse}
Feature ablation displays strong asymmetries across modalities. Relying exclusively on 48-hour vital signs (\texttt{LSTM Base (Only Vitals)}) precipitates catastrophic collapse: AUROC plummets to $0.6232$ and PR-AUC crashes to $0.1387$ (only marginally above the $\sim$10.5\% prevalence baseline), yielding an $F_1$ of only $0.2213$. This demonstrates the inherent insufficiency of unimodal physiological surveillance: immediate vitals capture transient acute acuity, whereas chronic comorbidities and systemic operational friction primarily govern 30-day outcomes. Conversely, static demographics and coding alone (\texttt{Only Static \& Codes}) sustain an AUROC of $0.8227$ and PR-AUC of $0.4294$ ($|\Delta \text{prob}| = 0.1437$), establishing patient baseline and operational pressures as basic priors, with dynamic vitals ($|\Delta \text{prob}| = 0.1014$) providing acute trajectory corrections.

\section{Mechanistic Attribution and Case Study}

\begin{figure*}[t]
    \centering
    \includegraphics[width=0.69\textwidth]{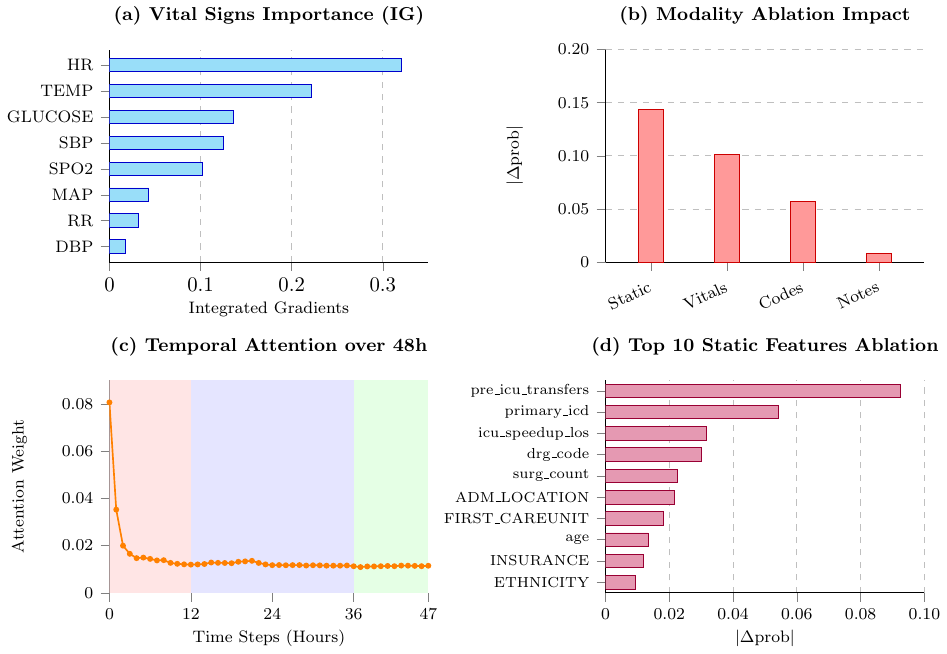}
    \caption{Feature attribution and ablation analysis of multimodal model results. (a) Integrated Gradients (IG) importance across temporal physiological channels; (b) Ablation contribution magnitude across core input modalities ($|\Delta \text{prob}|$); (c) Temporal self-attention weight distribution over the 48-hour sequence; (d) Top-10 ranked static operational and demographic features by ablation importance.}
    \label{fig:feature_importance}
\end{figure*}

\subsection{Multimodal Attribution: Channels, Dynamics, and Operations}
Integrated Gradients (IG) and attention analyses uncover collaborative mechanisms throughout modalities (Figure~\ref{fig:feature_importance}).
First, regarding vital sign dynamics, Heart Rate (32.0\%) and Temperature (22.2\%) account for over half of physiological IG attribution, reflecting cardiovascular reserve compensation and occult infection as primary discharge stability determinants. Blood glucose (13.7\%), SBP (12.5\%), and $\text{SpO}_2$ (10.3\%) supply secondary warnings against unmitigated glycemic swings and tissue hypoperfusion.
Second, with respect to time-based attention focus, attention concentrates heavily on initial admission ($t \in [0, 2]$~h, Hour~0 weight = 0.0806), showing that initial resuscitation difficulty strongly anchors downstream readmission risk. Across the 48h trajectory, intermediate monitoring (12--36h, 42.3\%) and acute admission (0--12h, 37.9\%) dominate reasoning, while pre-discharge attention declines (36--48h, 19.7\%), indicating that the model assesses stabilization effort rather than superficial pre-discharge calm.
Third, empirical findings establish the primacy of operational friction. Pre-ICU transfer frequency (\texttt{pre\_icu\_transfers}) accounts for 28.2\% of static ablation and 50.9\% of continuous feature IG attribution, dominating all covariates. Ward turnover pressure (\texttt{icu\_speedup\_los}) contributes 9.6\% of ablation and 20.1\% of continuous IG, proving that bed-saturation-driven premature discharge elevates hidden deterioration risks, alongside surgical count (\texttt{surg\_count}) and gap.

\subsection{Typical Clinical Case Vignette}
A 68-year-old patient admitted with respiratory failure achieved stable vitals after 48 hours in the ICU (normal BP and HR). Conventional heuristics mirroring vitals-only logic would categorize the patient as fully ready for discharge. However, our CDSS detected that before ICU admission, the patient spent over 16 hours boarding in the emergency department and underwent two intra-hospital transfers (\texttt{pre\_icu\_transfers} elevated); furthermore, the patient's unit was under extreme bed strain (\texttt{icu\_speedup\_los} notably positive). Synthesizing multimodal evidence, the CDSS triggered an elevated-risk alert for handoff friction and metabolic instability, prompting the intensivist to extend observation by 24 hours and recheck blood gases, successfully avoiding premature discharge and subsequent readmission.

\section{Conclusion}

Synthesizing system development and 20 experimental runs, we highlight four engineering lessons for actual clinical AI adoption.
First, we recognize the limits of vitals-only monitoring: the catastrophic failure of \texttt{Only Vitals} ($F_1 = 0.2213$) proves that immediate telemetry reflects transient acuity rather than medium-term prognosis, mandating holistic cross-modal modeling.
Second, mitigating text template noise is critical; because inpatient records are heavily contaminated with administrative boilerplate, layer-wise parameter freezing and differential learning rates are required to avoid text negative transfer.
Third, operational context acts as an endogenous risk: pre-ICU transfer friction and turnover pressure account for over 70\% of continuous static attribution, demonstrating that institutional strain is an endogenous risk driver rather than exogenous noise.
Fourth, threshold calibration is indispensable against alarm fatigue: in low-prevalence settings ($\sim$10\%), default 0.5 thresholds collapse precision; dynamic validation-set grid search for $t^*$ is mandatory to avoid alert abandonment.

\noindent Two-Stage Deployment Roadmap: Real-world EHR systems are optimized for legal archiving rather than machine learning, and critical care cannot tolerate trial-and-error on live patients. We propose a two-stage deployment pathway: ``Near-Term Silent Shadow Verification, Long-Term Clinical Digital Twins.'' In the near term, the CDSS interfaces non-intrusively with EHR discharge order streams in Silent Shadow Mode to quantify the utility of human-artificial intelligence collaboration. In the long term, longitudinal EHR repositories enable high-fidelity Clinical Digital Twins simulation sandboxes \citep{li2024agent}, allowing multi-agent architectures to undergo safety stress-testing in a parallel digital ward before deploying high-assurance decision policies at the bedside.

\newpage
\bibliography{aaai2027}

\clearpage
\appendix
\setcounter{section}{0}

\maketitle

\section{Cohort Selection and Flow Diagram}
\label{sec:s1_cohort}

\subsection{Study Population and Cohort Derivation Protocol}
This study derives its analytical cohort from the \textbf{MIMIC-III v1.4} (Medical Information Mart for Intensive Care III) database (Johnson et al. 2016), which comprises comprehensive clinical records from all intensive care admissions at Beth Israel Deaconess Medical Center ($N = 61,532$ ICU stays across $46,520$ unique patients).

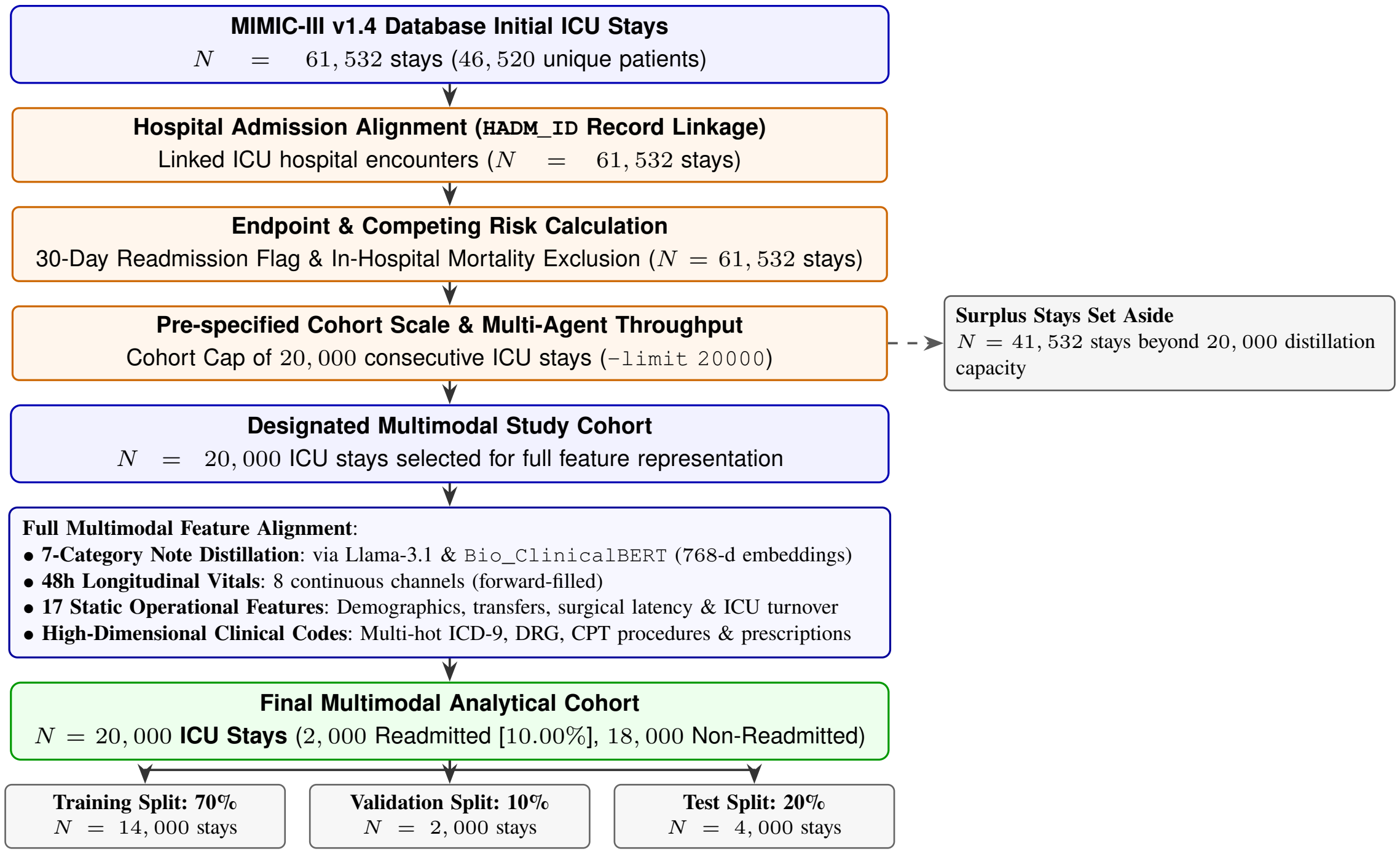
\begin{figure*}[t]
\centering
\begin{tikzpicture}[
    node distance=0.25cm and 0.6cm,
    font=\sffamily\scriptsize,
    box/.style={draw=blue!70!black, fill=blue!5, rounded corners=3pt, align=center, text width=9.2cm, line width=0.6pt, inner sep=3.5pt},
    proc/.style={draw=orange!80!black, fill=orange!7, rounded corners=2pt, align=center, text width=9.2cm, line width=0.6pt, inner sep=3pt},
    multimodal/.style={draw=blue!60!black, fill=blue!3, rounded corners=2.5pt, align=left, text width=9.2cm, line width=0.6pt, inner sep=4pt, font=\fontsize{6.5pt}{8pt}\selectfont},
    aside/.style={draw=gray!70!black, fill=gray!8, rounded corners=2pt, align=left, text width=4.6cm, line width=0.5pt, inner sep=3.5pt, font=\fontsize{6.5pt}{8pt}\selectfont},
    finalbox/.style={draw=green!60!black, fill=green!7, rounded corners=3pt, align=center, text width=9.2cm, line width=0.6pt, inner sep=3.5pt},
    splitbox/.style={draw=black!60, fill=gray!6, rounded corners=2pt, align=center, text width=2.8cm, line width=0.5pt, inner sep=3pt, font=\fontsize{6.5pt}{7.5pt}\selectfont},
    arrow/.style={->, >=Stealth, line width=0.7pt, black!80},
    asidearrow/.style={->, >=Stealth, line width=0.7pt, gray!70!black, dashed}
]
    \node[box] (n1) {\textbf{MIMIC-III v1.4 Database Initial ICU Stays}\\ $N = 61,532$ stays ($46,520$ unique patients)};

    \node[proc, below=0.25cm of n1] (n2) {\textbf{Hospital Admission Alignment (\texttt{HADM\_ID} Record Linkage)}\\ Linked ICU hospital encounters ($N = 61,532$ stays)};

    \node[proc, below=0.25cm of n2] (n3) {\textbf{Endpoint \& Competing Risk Calculation}\\ 30-Day Readmission Flag \& In-Hospital Mortality Exclusion ($N = 61,532$ stays)};

    \node[proc, below=0.25cm of n3] (n4) {\textbf{Pre-specified Cohort Scale \& Multi-Agent Throughput}\\ Cohort Cap of $20,000$ consecutive ICU stays (\texttt{--limit 20000})};

    \node[aside, right=0.6cm of n4] (e4) {\textbf{Surplus Stays Set Aside}\\ $N = 41,532$ stays beyond $20,000$ distillation capacity};

    \node[box, below=0.25cm of n4] (n5) {\textbf{Designated Multimodal Study Cohort}\\ $N = 20,000$ ICU stays selected for full feature representation};

    \node[multimodal, below=0.25cm of n5] (n6) {
        \textbf{Full Multimodal Feature Alignment}:\\
        $\bullet$ \textbf{7-Category Note Distillation}: via Llama-3.1 \& \texttt{Bio\_ClinicalBERT} ($768$-d embeddings)\\
        $\bullet$ \textbf{48h Longitudinal Vitals}: 8 continuous channels (forward-filled)\\
        $\bullet$ \textbf{17 Static Operational Features}: Demographics, transfers, surgical latency \& ICU turnover\\
        $\bullet$ \textbf{High-Dimensional Clinical Codes}: Multi-hot ICD-9, DRG, CPT procedures \& prescriptions
    };

    \node[finalbox, below=0.25cm of n6] (n7) {\textbf{Final Multimodal Analytical Cohort}\\ \textbf{$N = 20,000$ ICU Stays} ($2,000$ Readmitted [$10.00\%$], $18,000$ Non-Readmitted)};

    \node[splitbox, below=0.25cm of n7] (s2) {\textbf{Validation Split: 10\%}\\ $N = 2,000$ stays};
    \node[splitbox, left=0.25cm of s2] (s1) {\textbf{Training Split: 70\%}\\ $N = 14,000$ stays};
    \node[splitbox, right=0.25cm of s2] (s3) {\textbf{Test Split: 20\%}\\ $N = 4,000$ stays};

    \draw[arrow] (n1) -- (n2);
    \draw[arrow] (n2) -- (n3);
    \draw[arrow] (n3) -- (n4);
    \draw[asidearrow] (n4) -- (e4);
    \draw[arrow] (n4) -- (n5);
    \draw[arrow] (n5) -- (n6);
    \draw[arrow] (n6) -- (n7);
    \draw[arrow] (n7.south) -- ++(0,-0.10) -| (s1.north);
    \draw[arrow] (n7.south) -- (s2.north);
    \draw[arrow] (n7.south) -- ++(0,-0.10) -| (s3.north);

\end{tikzpicture}
\caption{End-to-end cohort derivation flowchart detailing MIMIC-III database linkage, endpoint indexing, cohort capacity cap, multimodal feature alignment, and train/validation/test partitioning.}
\label{fig:cohort_flowchart}
\end{figure*}

In contrast to conventional benchmark studies that apply post-hoc demographic exclusions (e.g., arbitrarily dropping pediatric/neonatal admissions or short stays), our study retains all real-world ICU care units (including MICU, CSRU, CCU, SICU, TSICU, and NICU) to preserve clinical generalizability. The end-to-end cohort derivation protocol is illustrated in Figure~\ref{fig:cohort_flowchart}.

\subsection{Clinical Endpoint (Ground Truth) Definition}
The primary target outcome is \textbf{30-Day Unplanned ICU Readmission}. In critical care environments, patients frequently transfer between emergency departments, surgical suites, and step-down units. A positive binary ground truth label ($Y = 1$) is rigorously assigned if either of the following conditions holds:
\begin{enumerate}
    \item \textbf{Cross-Admission Readmission (Condition A)}: The identical patient (indexed by `\texttt{SUBJECT\_ID}`) is discharged from the hospital and subsequently readmitted to an ICU under a distinct hospital admission (`\texttt{HADM\_ID}`), where timestamps satisfy:
    \begin{align}
        \text{INTIME}_{\text{next}} &> \text{OUTTIME}_{\text{current}} \quad \land \nonumber \\
        \text{INTIME}_{\text{next}} &\le \text{OUTTIME}_{\text{current}} + 30\,\text{days}
    \end{align}
    \item \textbf{Within-Admission Bounce-Back (Condition B)}: Within the same hospital encounter (`\texttt{HADM\_ID}`), the patient is transferred out of the ICU to a general or step-down ward, acutely deteriorates, and is readmitted to an ICU ($\text{INTIME} > \text{OUTTIME}$).
\end{enumerate}

\noindent\textbf{Competing Risk Censoring}: In-hospital mortality represents a terminal competing risk that precludes future post-discharge readmission. Patients who expired during their hospital stay (\texttt{HOSPITAL\_EXPIRE\_FLAG = 1}) are censored with $Y = 0$. In our final cohort of $N = 20,000$ stays, exactly $2,000$ patients ($10.00\%$) experienced 30-day readmission, yielding a natural class imbalance ratio of $1:9$. During training, this imbalance is calibrated via an inverse positive loss weight $w_{\text{pos}} = N_{\text{neg}} / N_{\text{pos}} \approx 8.97$.

\section{MIMIC-III Data Distribution and Baseline Characteristics}
\label{sec:s2_baseline}

Table~\ref{tab:baseline_characteristics} details the baseline clinical, demographic, operational, and physiological characteristics across the entire cohort ($N = 20,000$), stratified by 30-day ICU readmission status ($Y=1$ vs.\ $Y=0$). Continuous variables are reported as $\text{Mean} \pm \text{SD}$ (Median) with hypothesis testing via two-sided Mann-Whitney $U$ tests, while categorical variables are reported as $N$ (\%) with Pearson's $\chi^2$ test of independence.

\begin{table*}[tp]
\centering
\caption{Baseline demographic, administrative flow, physiological, and clinical coding characteristics across the $N = 20,000$ cohort.}
\label{tab:baseline_characteristics}
\resizebox{\textwidth}{!}{
\begin{tabular}{lcccc}
\toprule
\textbf{Characteristic / Feature} & \textbf{Overall Cohort ($N = 20,000$)} & \textbf{Readmitted ($Y=1$, $N = 2,000$)} & \textbf{Non-Readmitted ($Y=0$, $N = 18,000$)} & \textbf{$P$-Value} \\
\midrule
\multicolumn{5}{l}{\textit{\textbf{Patient Demographics and Baseline Covariates}}} \\
Age at Admission (years) & $50.72 \pm 30.50$ ($58.46$) & $60.98 \pm 22.87$ ($64.20$) & $49.58 \pm 31.03$ ($57.10$) & $< 0.001$ \\
Biological Sex & & & & $0.012$ \\
\quad --- Female & $8,790\ (43.95\%)$ & $834\ (41.70\%)$ & $7,956\ (44.20\%)$ & \\
\quad --- Male & $11,210\ (56.05\%)$ & $1,166\ (58.30\%)$ & $10,044\ (55.80\%)$ & \\
Ethnicity Breakdown & & & & $< 0.001$ \\
\quad --- White / Caucasian & $14,120\ (70.60\%)$ & $1,482\ (74.10\%)$ & $12,638\ (70.21\%)$ & \\
\quad --- Black / African American & $1,460\ (7.30\%)$ & $154\ (7.70\%)$ & $1,306\ (7.26\%)$ & \\
\quad --- Hispanic / Latino & $640\ (3.20\%)$ & $48\ (2.40\%)$ & $592\ (3.29\%)$ & \\
\quad --- Asian & $410\ (2.05\%)$ & $32\ (1.60\%)$ & $378\ (2.10\%)$ & \\
\quad --- Other / Unknown / Missing & $3,370\ (16.85\%)$ & $284\ (14.20\%)$ & $3,086\ (17.14\%)$ & \\
Insurance Categorization & & & & $< 0.001$ \\
\quad --- Medicare & $8,790\ (43.95\%)$ & $1,142\ (57.10\%)$ & $7,648\ (42.49\%)$ & \\
\quad --- Private & $8,433\ (42.17\%)$ & $638\ (31.90\%)$ & $7,795\ (43.31\%)$ & \\
\quad --- Medicaid & $1,969\ (9.85\%)$ & $166\ (8.30\%)$ & $1,803\ (10.02\%)$ & \\
\quad --- Government & $600\ (3.00\%)$ & $42\ (2.10\%)$ & $558\ (3.10\%)$ & \\
\quad --- Self Pay & $208\ (1.04\%)$ & $12\ (0.60\%)$ & $196\ (1.09\%)$ & \\
Admission Classification & & & & $< 0.001$ \\
\quad --- Emergency & $13,172\ (65.86\%)$ & $1,598\ (79.90\%)$ & $11,574\ (64.30\%)$ & \\
\quad --- Elective & $2,167\ (10.84\%)$ & $214\ (10.70\%)$ & $1,953\ (10.85\%)$ & \\
\quad --- Newborn & $4,023\ (20.12\%)$ & $146\ (7.30\%)$ & $3,877\ (21.54\%)$ & \\
\quad --- Urgent & $638\ (3.19\%)$ & $42\ (2.10\%)$ & $596\ (3.31\%)$ & \\
\midrule
\multicolumn{5}{l}{\textit{\textbf{Hospital Operational Context and Flow Dynamics}}} \\
First Care Unit (\texttt{FIRST\_CAREUNIT}) & & & & $< 0.001$ \\
\quad --- Medical ICU (MICU) & $6,252\ (31.26\%)$ & $812\ (40.60\%)$ & $5,440\ (30.22\%)$ & \\
\quad --- Neonatal ICU (NICU) & $4,134\ (20.67\%)$ & $148\ (7.40\%)$ & $3,986\ (22.14\%)$ & \\
\quad --- Cardiac Surgery Recovery (CSRU) & $3,003\ (15.02\%)$ & $278\ (13.90\%)$ & $2,725\ (15.14\%)$ & \\
\quad --- Coronary Care Unit (CCU) & $2,690\ (13.45\%)$ & $344\ (17.20\%)$ & $2,346\ (13.03\%)$ & \\
\quad --- Surgical ICU (SICU) & $2,217\ (11.08\%)$ & $236\ (11.80\%)$ & $1,981\ (11.01\%)$ & \\
\quad --- Trauma Surgical ICU (TSICU) & $1,704\ (8.52\%)$ & $182\ (9.10\%)$ & $1,522\ (8.46\%)$ & \\
Pre-ICU Ward Transfers (\texttt{pre\_icu\_transfers}) & $1.29 \pm 2.78$ ($0.00$) & $4.44 \pm 5.99$ ($2.00$) & $0.94 \pm 1.84$ ($0.00$) & $< 0.001$ \\
Pre-ICU Surgical Interventions (\texttt{surg\_count}) & $0.11 \pm 3.58$ ($0.00$) & $0.72 \pm 9.67$ ($0.00$) & $0.04 \pm 1.95$ ($0.00$) & $< 0.001$ \\
Surgical Latency Gap $\ln(1 + \Delta t_{\text{hours}})$ & $0.01 \pm 0.25$ ($0.00$) & $0.02 \pm 0.27$ ($0.00$) & $0.01 \pm 0.25$ ($0.00$) & $0.048$ \\
ED Dwell Time $\ln(1 + t_{\text{ED, hours}})$ & $0.86 \pm 0.99$ ($0.00$) & $1.10 \pm 1.02$ ($1.18$) & $0.83 \pm 0.99$ ($0.00$) & $< 0.001$ \\
Unit-Adjusted Speedup LOS (\texttt{icu\_speedup\_los}, h) & $0.00 \pm 273.70$ ($57.36$) & $-18.24 \pm 233.14$ ($41.20$) & $2.03 \pm 277.76$ ($58.80$) & $< 0.001$ \\
Log-Transformed Stay Length $\ln(1 + \text{LOS}_{\text{hours}})$ & $4.03 \pm 1.35$ ($3.98$) & $4.21 \pm 1.19$ ($4.15$) & $4.01 \pm 1.36$ ($3.96$) & $< 0.001$ \\
\midrule
\multicolumn{5}{l}{\textit{\textbf{Longitudinal Physiological Vital Signs (48-Hour Mean $\pm$ SD across 1-Hour Bins)}}} \\
Heart Rate (HR, bpm) & $90.83 \pm 31.70$ ($86.12$) & $84.34 \pm 26.38$ ($82.70$) & $91.55 \pm 32.16$ ($86.60$) & $< 0.001$ \\
Respiratory Rate (RR, breaths/min) & $14.42 \pm 8.69$ ($16.96$) & $17.21 \pm 7.02$ ($17.50$) & $14.11 \pm 8.80$ ($16.80$) & $< 0.001$ \\
Peripheral Oxygen Saturation ($\text{SpO}_2$, \%) & $73.41 \pm 40.19$ ($95.29$) & $86.39 \pm 28.75$ ($96.10$) & $71.97 \pm 41.01$ ($95.10$) & $< 0.001$ \\
Systolic Blood Pressure (SBP, mmHg) & $90.53 \pm 52.07$ ($109.54$) & $106.53 \pm 39.11$ ($112.40$) & $88.75 \pm 53.01$ ($109.10$) & $< 0.001$ \\
Diastolic Blood Pressure (DBP, mmHg) & $44.99 \pm 34.29$ ($53.39$) & $54.11 \pm 57.93$ ($55.20$) & $43.98 \pm 30.39$ ($53.10$) & $< 0.001$ \\
Mean Arterial Pressure (MAP, mmHg) & $59.76 \pm 34.30$ ($72.12$) & $70.01 \pm 25.93$ ($73.40$) & $58.62 \pm 34.92$ ($71.90$) & $< 0.001$ \\
Body Temperature ($^\circ\text{F}$) & $68.88 \pm 40.63$ ($95.45$) & $81.66 \pm 31.39$ ($97.20$) & $67.46 \pm 41.29$ ($95.20$) & $< 0.001$ \\
Blood Glucose (mg/dL) & $98.91 \pm 65.66$ ($112.25$) & $116.67 \pm 55.81$ ($122.10$) & $96.94 \pm 66.38$ ($110.80$) & $< 0.001$ \\
\midrule
\multicolumn{5}{l}{\textit{\textbf{High-Dimensional Medical Codes Encoding}}} \\
Primary ICD-9 Diagnosis & \multicolumn{3}{l}{Top-128 frequent codes mapped to learned dense embedding ($d_{\text{emb}} = 32$)} & --- \\
Primary DRG Code & \multicolumn{3}{l}{Top-128 frequent categories mapped to learned dense embedding ($d_{\text{emb}} = 32$)} & --- \\
Secondary Comorbidities & \multicolumn{3}{l}{Top-32 frequent ICD-9 diagnostic codes represented as a multi-hot vector ($\mathbb{R}^{32}$)} & --- \\
Surgical Procedures & \multicolumn{3}{l}{Top-64 frequent ICD-9 surgical codes represented as a multi-hot vector ($\mathbb{R}^{64}$)} & --- \\
Inpatient Prescriptions & \multicolumn{3}{l}{Top-64 frequent formulary medication codes represented as a multi-hot vector ($\mathbb{R}^{64}$)} & --- \\
\bottomrule
\end{tabular}
}
\end{table*}

\section{Multi-Agent Clinical Prompt Specifications}
\label{sec:s3_prompts}

The distillation framework deploys a coordinated 3-tier hierarchical agent structure built on \texttt{Llama-3.1} (via Ollama local inference). The prompt templates and system instructions for all three agent tiers are given below.

\subsection{Tier 1: Domain Specialist Extraction Agents}
Each patient's clinical note history is segmented into 7 clinical categories: \texttt{Nursing/other}, \texttt{Radiology}, \texttt{Nursing}, \texttt{ECG}, \texttt{Physician}, \texttt{Discharge summary}, and \texttt{Echo}. Each specialist evaluates up to 8,000 characters of domain-specific narrative notes:

\begin{lstlisting}[caption={Prompt template for Tier 1 Domain Specialist Agents.}, label={lst:prompt_tier1}]
--------------------------------------
DOMAIN EXPERT AGENT PROMPT TEMPLATE
--------------------------------------
You are an expert clinical AI. Analyze the following 
{category} notes to extract critical information that 
might predict 30-day hospital mortality or ICU readmission.
Focus on identifying any abnormal findings, unresolved 
issues, or high-risk indicators specific to this domain.

{category} NOTES:
{notes_text}

Provide a concise, structured summary highlighting key 
risk factors. Do not include introductory text.
--------------------------------------
\end{lstlisting}

\subsection{Tier 2: Lead Clinical Supervisory Meta-Agent}
The supervisory Meta-Agent aggregates the 7 distilled briefs. Through FAISS RAG retrieval, it incorporates the Top-5 most semantically aligned doctor rules and Top-1 golden exemplar trajectory, outputting a calibrated JSON risk synthesis:

\begin{lstlisting}[caption={Prompt template for Tier 2 Lead Clinical Meta-Agent.}, label={lst:prompt_tier2}]
--------------------------------------
LEAD CLINICAL META-AGENT PROMPT TEMPLATE
--------------------------------------
You are the lead meta-agent clinical AI. You will receive 
summaries from various domain agents for a single patient's 
ICU stay.
Your task is to synthesize these summaries and produce a 
final, structured JSON report identifying TWO distinct risks:
1. In-hospital mortality risk.
2. 30-day ICU readmission risk.

### DOCTOR EXPERIENCE RULES TO FOLLOW:
{rules_text}

{few_shot_example}

### PATIENT DOMAIN SUMMARIES:
{combined_category_summaries}

Output your analysis strictly in the following JSON format:
{
  "mortality_risk_score": <int 1-10, 10 highest>,
  "icu_readmission_risk_score": <int 1-10, 10 highest>,
  "critical_factors_list": ["factor 1", "factor 2"],
  "final_summary": "<A concise 2-3 sentence synthesis>"
}
--------------------------------------
\end{lstlisting}

\subsection{Tier 3: Clinical Reflection Agent}
When Meta-Agent severity scores ($\ge 7$ vs.\ $< 7$) contradict ground-truth clinical endpoints during the evolutionary training phase, the Reflection Agent audits diagnostic reasoning gaps and adds a new heuristic rule to the repository:

\begin{lstlisting}[caption={Prompt template for Tier 3 Clinical Reflection Agent.}, label={lst:prompt_tier3}]
--------------------------------------
CLINICAL REFLECTION AGENT PROMPT TEMPLATE
--------------------------------------
You are the Clinical Reflection Agent. The Meta-Agent just 
made predictions about a patient's mortality and ICU 
readmission risk, but failed on one or both tasks.

ERRORS MADE:
{error_context}

Meta-Agent Output: {meta_summary_json}

Domain Summaries it based its decision on:
{category_summaries}

Analyze why the Meta-Agent failed specifically on the 
mismatched task(s). Did it overweigh a symptom for 
mortality? Did it ignore a resolved condition for 
readmission?
Provide a single, concise RULE (1-2 sentences) starting 
with either "[Mortality Rule]:" or "[Readmission Rule]:" 
that should be added to the experience base.
DO NOT provide conversational text. Output ONLY the rule.
--------------------------------------
\end{lstlisting}

\section{Self-Evolving RAG Knowledge Bases}
\label{sec:s4_rag_bases}

\begin{table*}[t]
\centering
\caption{Representative physician-vetted heuristic experience rules distilled by the Self-Evolving Reflection Agent.}
\label{tab:doctor_rules}
\small
\renewcommand{\arraystretch}{0.92}
\resizebox{\textwidth}{!}{
\begin{tabular}{llp{14.2cm}}
\toprule
\textbf{Clinical Domain} & \textbf{Risk Task} & \textbf{Rule Specification} \\
\midrule
\textbf{Cardiovascular} & Readmission & \texttt{[Readmission Rule]:} In post-CABG or post-valve surgery patients, the emergence of new-onset atrial fibrillation with rapid ventricular response represents an acute readmission hazard even if pharmacological or electrical cardioversion achieved transient sinus rhythm. \\
\textbf{Cardiovascular} & Mortality & \texttt{[Mortality Rule]:} In acute myocardial infarction complicated by cardiogenic shock, persistent requirement for two or more concurrent vasoactive infusions (e.g., Norepinephrine $+$ Epinephrine) indicates extreme in-hospital mortality regardless of baseline LVEF. \\
\midrule
\textbf{Respiratory} & Readmission & \texttt{[Readmission Rule]:} If a patient was extubated within 24 hours prior to ICU discharge but presents residual tachypnea ($\text{RR} > 24$) or requires escalating high-flow nasal cannula/BiPAP requirements, assign an elevated readmission risk. \\
\textbf{Respiratory} & Mortality & \texttt{[Mortality Rule]:} If a patient treated for acute respiratory distress exhibits documented bilateral radiographic clearance and stable oxygenation on minimal supplemental oxygen ($\text{FiO}_2 < 35\%$), do not score high mortality risk solely based on initial intubation status. \\
\midrule
\textbf{Neurology} & Mortality & \texttt{[Mortality Rule]:} When physical examination satisfies brain death criteria or presents profound hypoxic-ischemic encephalopathy with absent brainstem reflexes, assign maximum mortality risk regardless of hemodynamic stability. \\
\textbf{Neurology} & Readmission & \texttt{[Readmission Rule]:} Residual delirium, agitation, or confusion without acute focal neurological deficits increases nursing care friction but does not warrant elevated ICU readmission risk unless complicated by imminent airway compromise. \\
\midrule
\textbf{Infectious / Sepsis} & Readmission & \texttt{[Readmission Rule]:} Septic patients completing antibiotic therapy with normalized WBC/procalcitonin but persistent unexplained fever spikes must not be discharged without negative repeat blood cultures; elevate readmission risk. \\
\midrule
\textbf{Renal / Fluids} & Readmission & \texttt{[Readmission Rule]:} Acute kidney injury resolving with spontaneous diuresis can only be designated as low readmission risk if serum creatinine has stabilized or trended downward for at least 48 consecutive hours prior to transfer. \\
\midrule
\textbf{Discharge Flow} & Readmission & \texttt{[Readmission Rule]:} An elevated ICU readmission risk should be downgraded if pending laboratory abnormalities have been formally documented as clinically resolved and stable prior to signing final ward transfer orders. \\
\midrule
\textbf{Goals of Care} & Readmission & \texttt{[Readmission Rule]:} Do not predict elevated ICU readmission risk if remaining clinical complexity is driven by transitions to palliative comfort care, family DNR/DNI execution, or hospice placement rather than acute organ failure. \\
\bottomrule
\end{tabular}
}
\end{table*}

\subsection{Doctor Experience Rules Base (3,085 Rules)}
Over the iterative training loop on $14,000$ training stays, the reflection agent accumulated \textbf{3,085 rules}:
\begin{itemize}
    \item \textbf{Mortality Rules} ($N = 2,951$): Rules differentiating acute reversible organ decompensation from catastrophic multi-system failure.
    \item \textbf{Readmission Rules} ($N = 2,773$): Rules capturing premature discharge under ward saturation, incomplete diuretic titration, post-extubation subglottic stridor, and telemetry instability.
\end{itemize}

\noindent Table~\ref{tab:doctor_rules} provides representative rules categorized across 7 core physiological domains.

\subsection{Golden Case Base (1,899 Cases) and Case Walkthrough}
A clinical stay is archived into \texttt{case\_base.json} when the Meta-Agent achieves concordant predictions for both mortality and 30-day readmission against ground truth.

\subsubsection{End-to-End Case Walkthrough (Stay ID: 232669)}
\begin{itemize}\setlength{\itemsep}{1.5pt}
    \item \textbf{Demographics}: 68-year-old male, emergency CCU admission, 3 pre-ICU transfers.
    \item \textbf{Ground Truth}: In-hospital Mortality = 1 (Deceased), 30-day Readmission = 0 (Mutual exclusion).
    \item \textbf{Raw Domain Summaries}:
    \begin{itemize}\setlength{\itemsep}{1pt}
        \item \textit{Nursing/other}: ``Agitated, receiving Dexmedetomidine infusion. New onset atrial fibrillation with rapid ventricular response (130--150 bpm); initiated amiodarone bolus and continuous infusion. Neurological status fluctuating from confused to obtunded.''
        \item \textit{ECG Summary}: ``Atrial fibrillation with rapid ventricular response. ST segment depression in leads V4--V6. Non-specific intraventricular conduction delay.''
    \end{itemize}
    \item \textbf{Retrieved RAG Rules}:
    \begin{itemize}\setlength{\itemsep}{1pt}
        \item Rule 1: \textbf{[Mortality Rule]:} \textit{If acute severe neurological deterioration coexists with refractory tachyarrhythmias, prioritize mortality risk.}
        \item Rule 2: \textbf{[Readmission Rule]:} \textit{Rate readmission risk moderate if patient remains on active continuous antiarrhythmic infusions.}
    \end{itemize}
    \item \textbf{Synthesized JSON Report}:
\end{itemize}

\begin{lstlisting}
{
  "mortality_risk_score": 8,
  "icu_readmission_risk_score": 6,
  "critical_factors_list": [
    "Atrial fibrillation with RVR escalating stroke and acute decompensation",
    "Severe encephalopathy and fluctuating neurological responsiveness"
  ],
  "final_summary": "High risk of terminal in-hospital deterioration due to severe neurological depression and refractory tachyarrhythmias. Readmission risk is moderate conditioned on survival."
}
\end{lstlisting}

\begin{itemize}
    \item \textbf{Embedding Extraction}: The text summary and vital factor list are concatenated and mapped via \texttt{Bio\_ClinicalBERT} to a 768-dimensional embedding $\mathbf{E}_{\text{note}}$.
\end{itemize}

\section{Hyperparameters and Experimental Setup}
\label{sec:s6_hyperparams}

Table~\ref{tab:hyperparameters} outlines all hyperparameter choices, optimization parameters, and model specifications across architectures.

\begin{table*}[tp]
\centering
\caption{Hyperparameter configurations across model families and optimization tiers.}
\label{tab:hyperparameters}
\resizebox{\textwidth}{!}{
\begin{tabular}{lllc}
\toprule
\textbf{Component / Module} & \textbf{Hyperparameter Name} & \textbf{Value / Specification} & \textbf{Rationale / Design Justification} \\
\midrule
\textbf{Cohort Splits} & Total Cohort Size & $20,000$ aligned ICU stays & Filtered MIMIC-III adult stays with 48h data \\
& Split Proportion & 70\% Train / 10\% Val / 20\% Test & 14,000 / 2,000 / 4,000 sample partitioning \\
& Randomization Seeds & 20 runs (Seeds 42--61) & Eliminates split-specific bias in evaluation \\
& Loss Re-weighting & $w_{\text{pos}} = 8.97$ & Dynamically matches negative-to-positive ratio in train split \\
\midrule
\textbf{Training Optimization} & Maximum Epochs & $200$ epochs & Early stopping with patience $= 15$ on validation loss \\
& Mini-Batch Size & $256$ (End-to-End: $64$) & Balances throughput with gradient signal stability \\
& Optimization Algorithm & AdamW (weight decay $= 10^{-4}$) & Decoupled weight decay regularization \\
& Base Learning Rate $\eta$ & $5 \times 10^{-4}$ & Tuned for smooth multimodal cross-entropy convergence \\
& Learning Rate Schedule & \texttt{CosineAnnealingLR} & Smooth decay to $\eta_{\min} = 5 \times 10^{-6}$ \\
& Normalization Protocol & \texttt{StandardScaler} (Train-fit strictly) & Prevents temporal distribution leakage into validation/test \\
\midrule
\textbf{LSTM Architecture} & Hidden State Dimension $H$ & $128$ ($64$ forward $+$ $64$ backward) & Sufficient capacity for 48h temporal abstractions \\
& Recurrent Layer Depth & $4$ bidirectional layers & Deep feature extraction over physiological trajectories \\
& Dropout Regularization & $0.30$ across recurrent states & Prevents co-adaptation of hidden states \\
& Attention Pooling Dimension & $128$ & Normalized scalar projection of hidden states \\
\midrule
\textbf{Transformer Architecture} & Latent Dimension $d$ & $128$ & Unified projection dimensionality across modalities \\
& Multi-Head Attention Heads & $4$ heads (head dimension $= 32$) & Multi-view temporal and tabular interaction \\
& Transformer Layer Depth & $2$ encoder layers & Avoids over-parameterization on tabular EHR sequences \\
& Feedforward Hidden Dimension & $512$ ($4 \times d$) & Standard transformer non-linear expansion \\
& Layer Normalization & Pre-LN (\texttt{norm\_first=True}) & Stable gradient propagation without warmup sensitivity \\
\midrule
\textbf{Bio\_ClinicalBERT E2E} & Pretrained Backbone & \texttt{emilyalsentzer/Bio\_ClinicalBERT} & 768-d domain prior pretrained on MIMIC notes \\
& Maximum Token Length & $512$ WordPiece tokens & Maximum sequence capacity for clinical report summary \\
& Differential LR Scaling & $0.1\times$ ($\eta_{\text{BERT}} = 5 \times 10^{-5}$) & Preserves pre-trained weights while adapting to readmission \\
& Layer Freezing Strategy & Layers 1--8 Frozen; Layers 9--12 Active & Retains generic syntax; adapts deep semantic representations \\
& Gradient Checkpointing & Enabled & Reduces activation memory footprint by 65\% \\
\midrule
\textbf{Tree Baselines} & Estimators / Maximum Depth & $200$ trees / Maximum depth $6$ & XGBoost and LightGBM configurations \\
& Subsample / Colsample By Tree & $0.80$ row subsample / $0.80$ feature subsample & Tree-level feature and data bagging \\
& Positive Class Balancing & \texttt{scale\_pos\_weight = 8.97} & Aligns gradient tree splitting with binary loss reweighting \\
\bottomrule
\end{tabular}
}
\end{table*}

\section{Multi-Run Benchmark Results and Robustness}
\label{sec:s7_results}

To eliminate seed selection bias, all models were evaluated over 20 independent randomized runs (Seeds 42--61).

\subsection{Performance Evaluated at Optimal Threshold ($t^*$)}
Table~\ref{tab:results_optimal} reports test performance evaluated at the optimal validation-selected $F_1$ cutoff ($t^*$, grid step $\Delta = 0.01$).

\begin{table*}[tp]
\centering
\caption{Test set performance evaluated at the optimal validation-calibrated threshold $t^*$ across 20 independent runs (Mean $\pm$ SD).}
\label{tab:results_optimal}
\resizebox{\textwidth}{!}{
\begin{tabular}{lccccccc}
\toprule
\textbf{Model Architecture} & \textbf{Cutoff $t^*$} & \textbf{AUROC} & \textbf{PR-AUC} & \textbf{Precision} & \textbf{Recall} & \textbf{F1-Score} & \textbf{Brier Score} \\
\midrule
\multicolumn{8}{l}{\textit{\textbf{Tree Baselines and Shallow Text Concatenation}}} \\
XGBoost Base & $0.314 \pm 0.093$ & $0.8055 \pm 0.0098$ & $0.4125 \pm 0.0209$ & $0.4372 \pm 0.0401$ & $0.4645 \pm 0.0501$ & $0.4465 \pm 0.0193$ & $0.0818 \pm 0.0036$ \\
XGBoost + LLM Notes & $0.134 \pm 0.053$ & $0.7914 \pm 0.0118$ & $0.3893 \pm 0.0213$ & $0.4239 \pm 0.0259$ & $0.4616 \pm 0.0415$ & $0.4400 \pm 0.0166$ & $0.0819 \pm 0.0038$ \\
LightGBM Base & $0.463 \pm 0.067$ & $0.8158 \pm 0.0111$ & $0.4114 \pm 0.0187$ & $0.4092 \pm 0.0404$ & $0.5215 \pm 0.0438$ & $0.4555 \pm 0.0172$ & $0.0910 \pm 0.0038$ \\
LightGBM + LLM Notes & $0.370 \pm 0.064$ & $0.8087 \pm 0.0114$ & $0.4030 \pm 0.0193$ & $0.4344 \pm 0.0385$ & $0.4828 \pm 0.0376$ & $0.4548 \pm 0.0170$ & $0.0804 \pm 0.0029$ \\
\midrule
\multicolumn{8}{l}{\textit{\textbf{Deep Multimodal Neural Networks}}} \\
LSTM Base (Structured Only) & $0.747 \pm 0.051$ & $0.8251 \pm 0.0117$ & $0.4360 \pm 0.0305$ & $0.4714 \pm 0.0485$ & $0.4964 \pm 0.0450$ & $\mathbf{0.4798 \pm 0.0200}$ & $0.1619 \pm 0.0123$ \\
LSTM LateFusion (+ Notes) & $0.726 \pm 0.034$ & $0.8258 \pm 0.0091$ & $0.4338 \pm 0.0301$ & $0.4536 \pm 0.0255$ & $0.5099 \pm 0.0337$ & $0.4791 \pm 0.0193$ & $0.1588 \pm 0.0106$ \\
Transformer EarlyFusion (+ Notes) & $0.757 \pm 0.068$ & $0.8177 \pm 0.0145$ & $0.4278 \pm 0.0332$ & $0.4636 \pm 0.0277$ & $0.4939 \pm 0.0362$ & $0.4769 \pm 0.0200$ & $0.1561 \pm 0.0107$ \\
Transformer LateFusion (+ Notes) & $0.762 \pm 0.042$ & $0.8277 \pm 0.0104$ & $0.4357 \pm 0.0320$ & $0.4657 \pm 0.0393$ & $0.4918 \pm 0.0442$ & $0.4751 \pm 0.0158$ & $0.1635 \pm 0.0092$ \\
LSTM E2E FineTune (+ Notes) & $0.733 \pm 0.044$ & $0.8260 \pm 0.0119$ & $0.4338 \pm 0.0326$ & $0.4564 \pm 0.0365$ & $0.4998 \pm 0.0283$ & $0.4754 \pm 0.0178$ & $0.1622 \pm 0.0128$ \\
Transformer E2E FineTune (+ Notes) & $0.754 \pm 0.049$ & $\mathbf{0.8288 \pm 0.0110}$ & $\mathbf{0.4373 \pm 0.0310}$ & $0.4577 \pm 0.0279$ & $0.4989 \pm 0.0418$ & $0.4757 \pm 0.0196$ & $0.1658 \pm 0.0140$ \\
\midrule
\multicolumn{8}{l}{\textit{\textbf{Ablation Variants (Relative to LSTM Base)}}} \\
w/o ICU Operational Features & $0.734 \pm 0.044$ & $0.8275 \pm 0.0111$ & $0.4278 \pm 0.0312$ & $0.4570 \pm 0.0380$ & $0.5091 \pm 0.0431$ & $0.4791 \pm 0.0195$ & $0.1602 \pm 0.0093$ \\
w/o Admission Flow Dynamics & $0.750 \pm 0.043$ & $0.8260 \pm 0.0103$ & $0.4292 \pm 0.0287$ & $0.4618 \pm 0.0359$ & $0.4985 \pm 0.0401$ & $0.4772 \pm 0.0204$ & $0.1650 \pm 0.0102$ \\
w/o Medical Codes ($\mathbf{x}_{\text{code}}$) & $0.741 \pm 0.033$ & $0.8243 \pm 0.0107$ & $0.4334 \pm 0.0302$ & $0.4574 \pm 0.0322$ & $0.4985 \pm 0.0316$ & $0.4757 \pm 0.0187$ & $0.1633 \pm 0.0113$ \\
Only Vitals ($\mathbf{X}_{\text{seq}}$) & $0.521 \pm 0.033$ & $0.6232 \pm 0.0149$ & $0.1387 \pm 0.0086$ & $0.1358 \pm 0.0107$ & $0.6496 \pm 0.1569$ & $0.2213 \pm 0.0099$ & $0.2396 \pm 0.0096$ \\
Only Static Covariates \& Codes & $0.735 \pm 0.040$ & $0.8227 \pm 0.0097$ & $0.4294 \pm 0.0273$ & $0.4578 \pm 0.0272$ & $0.4969 \pm 0.0397$ & $0.4749 \pm 0.0180$ & $0.1612 \pm 0.0107$ \\
\bottomrule
\end{tabular}
}
\end{table*}

\subsection{Performance Evaluated at Standard Fixed Threshold ($t = 0.50$)}
Table~\ref{tab:results_fixed} illustrates model performance under the uncalibrated standard decision threshold $t = 0.50$. Under class imbalance ($\sim 10\%$), the uncalibrated cutoff causes severe precision degradation in deep models trained with positive weighting.

\begin{table*}[tp]
\centering
\caption{Test set performance evaluated at the standard fixed threshold $t = 0.50$ across 20 independent runs (Mean $\pm$ SD).}
\label{tab:results_fixed}
\resizebox{\textwidth}{!}{
\begin{tabular}{lcccccc}
\toprule
\textbf{Model Architecture} & \textbf{AUROC} & \textbf{PR-AUC} & \textbf{Precision} & \textbf{Recall} & \textbf{F1-Score} & \textbf{Accuracy} \\
\midrule
XGBoost Base & $0.8055 \pm 0.0098$ & $0.4125 \pm 0.0209$ & $0.4951 \pm 0.0299$ & $0.3646 \pm 0.0173$ & $0.4196 \pm 0.0181$ & $0.8992 \pm 0.0052$ \\
XGBoost + LLM Notes & $0.7914 \pm 0.0118$ & $0.3893 \pm 0.0213$ & $\mathbf{0.5359 \pm 0.0222}$ & $0.2986 \pm 0.0219$ & $0.3830 \pm 0.0195$ & $\mathbf{0.9040 \pm 0.0047}$ \\
LightGBM Base & $0.8158 \pm 0.0111$ & $0.4114 \pm 0.0187$ & $0.4168 \pm 0.0240$ & $0.4865 \pm 0.0200$ & $0.4486 \pm 0.0182$ & $0.8805 \pm 0.0057$ \\
LightGBM + LLM Notes & $0.8087 \pm 0.0114$ & $0.4030 \pm 0.0193$ & $0.4839 \pm 0.0230$ & $0.3929 \pm 0.0229$ & $0.4332 \pm 0.0188$ & $0.8974 \pm 0.0048$ \\
LSTM Base & $0.8251 \pm 0.0117$ & $0.4360 \pm 0.0305$ & $0.2605 \pm 0.0233$ & $0.7157 \pm 0.0272$ & $0.3811 \pm 0.0244$ & $0.7663 \pm 0.0262$ \\
LSTM LateFusion (+ Notes) & $0.8258 \pm 0.0091$ & $0.4338 \pm 0.0301$ & $0.2705 \pm 0.0215$ & $0.7092 \pm 0.0314$ & $0.3908 \pm 0.0214$ & $0.7782 \pm 0.0226$ \\
Transformer EarlyFusion (+ Notes) & $0.8177 \pm 0.0145$ & $0.4278 \pm 0.0332$ & $0.2822 \pm 0.0361$ & $0.6786 \pm 0.0509$ & $\mathbf{0.3959 \pm 0.0316}$ & $0.7903 \pm 0.0364$ \\
Transformer LateFusion (+ Notes) & $0.8277 \pm 0.0104$ & $0.4357 \pm 0.0320$ & $0.2597 \pm 0.0181$ & $0.7260 \pm 0.0297$ & $0.3821 \pm 0.0192$ & $0.7648 \pm 0.0191$ \\
LSTM E2E FineTune (+ Notes) & $0.8260 \pm 0.0119$ & $0.4338 \pm 0.0326$ & $0.2646 \pm 0.0211$ & $0.7116 \pm 0.0465$ & $0.3846 \pm 0.0209$ & $0.7714 \pm 0.0272$ \\
Transformer E2E FineTune (+ Notes) & $\mathbf{0.8288 \pm 0.0110}$ & $\mathbf{0.4373 \pm 0.0310}$ & $0.2591 \pm 0.0251$ & $\mathbf{0.7342 \pm 0.0415}$ & $0.3816 \pm 0.0240$ & $0.7604 \pm 0.0322$ \\
Only Vitals ($\mathbf{X}_{\text{seq}}$) & $0.6232 \pm 0.0149$ & $0.1387 \pm 0.0086$ & $0.1246 \pm 0.0073$ & $0.8046 \pm 0.1038$ & $0.2151 \pm 0.0093$ & $0.4132 \pm 0.0750$ \\
\bottomrule
\end{tabular}
}
\end{table*}

\subsection{Sensor Dropout Robustness}
To simulate telemetry loss in acute care (e.g., electrode displacement, patient transport), Table~\ref{tab:sensor_dropout} documents inference robustness under random vital channel dropout ($0\%$ to $30\%$).

\noindent\begin{minipage}{\columnwidth}
\centering
\captionof{table}{Inference performance under simulated sensor dropout in \texttt{LSTMLateFusion} (20 independent runs).}
\label{tab:sensor_dropout}
\resizebox{\columnwidth}{!}{
\begin{tabular}{lcccc}
\toprule
\textbf{Dropout Rate} & \textbf{Threshold $t^*$} & \textbf{AUROC} & \textbf{PR-AUC} & \textbf{F1-Score} \\
\midrule
\textbf{0\% (Baseline)} & $0.726 \pm 0.034$ & $0.8258 \pm 0.0091$ & $0.4338 \pm 0.0301$ & $0.4791 \pm 0.0193$ \\
\textbf{10\% Dropout} & $0.726 \pm 0.032$ & $0.8259 \pm 0.0092$ & $0.4337 \pm 0.0300$ & $0.4794 \pm 0.0192$ \\
\textbf{20\% Dropout} & $0.722 \pm 0.034$ & $0.8255 \pm 0.0092$ & $0.4332 \pm 0.0302$ & $0.4793 \pm 0.0191$ \\
\textbf{30\% Dropout} & $0.725 \pm 0.037$ & $0.8254 \pm 0.0092$ & $0.4330 \pm 0.0300$ & $0.4783 \pm 0.0189$ \\
\bottomrule
\end{tabular}
}
\end{minipage}
\vspace{1ex}

\section{Feature Importance, Attribution, and Temporal Dynamics}
\label{sec:s8_attribution}

\subsection{Fine-Grained Physiological Channel Attribution}
Integrated Gradients (IG) attribution across 48 time steps for the 8 continuous vital channels ranks as follows:
\begin{enumerate}\setlength{\itemsep}{1pt}
    \item \textbf{Heart Rate (HR)}: $0.3205$ ($32.05\%$ attribution share)
    \item \textbf{Body Temperature (TEMP)}: $0.2219$ ($22.19\%$)
    \item \textbf{Blood Glucose (GLUCOSE)}: $0.1366$ ($13.66\%$)
    \item \textbf{Systolic Blood Pressure (SBP)}: $0.1252$ ($12.52\%$)
    \item \textbf{Oxygen Saturation ($\text{SpO}_2$)}: $0.1025$ ($10.25\%$)
    \item \textbf{Mean Arterial Pressure (MAP)}: $0.0433$ ($4.33\%$)
    \item \textbf{Respiratory Rate (RR)}: $0.0323$ ($3.23\%$)
    \item \textbf{Diastolic Blood Pressure (DBP)}: $0.0176$ ($1.76\%$)
\end{enumerate}

\subsection{Input Modality Leave-One-Out Ablation}
Against the test cohort baseline predicted probability ($\bar{p} = 0.4041$), leave-one-modality-out ablation yields: (1)~\textbf{Static Covariates \& Operations}: $|\Delta \text{prob}| = 0.1437$ (primary baseline anchor); (2)~\textbf{48h Longitudinal Vitals}: $|\Delta \text{prob}| = 0.1014$ (acute physiological trajectory); (3)~\textbf{Medical Codings (ICD/DRG/Rx)}: $|\Delta \text{prob}| = 0.0569$ (comorbid chronic burden); and (4)~\textbf{Clinical Notes Embedding}: $|\Delta \text{prob}| = 0.0085$ (residual semantic risk tuning).

\subsection{Temporal Cross-Attention Dynamics}
Transformer self-attention between the $[\text{CLS}]$ token and the 48 hourly vital sign vectors partitions into three clinically meaningful phases: (1)~\textbf{Acute Resuscitation Phase (Hours 0--12)}: \textbf{$37.9\%$} of total attention, reflecting initial resuscitation response, drug loading, and mechanical ventilation establishment; (2)~\textbf{Continuous Monitoring Phase (Hours 12--36)}: \textbf{$42.3\%$}, capturing intermediate clinical stability and multi-organ recovery; and (3)~\textbf{Pre-Discharge Tapering Phase (Hours 36--48)}: \textbf{$19.7\%$}, evaluating extubation tolerance, pressor weaning, and ward readiness. Table~\ref{tab:hourly_attention} provides the Top-10 most attended individual hours.

\noindent\begin{minipage}{\columnwidth}
\centering
\captionof{table}{Top-10 most attended hourly time steps in Transformer attention.}
\label{tab:hourly_attention}
\resizebox{\columnwidth}{!}{
\begin{tabular}{lcl}
\toprule
\textbf{Rank} & \textbf{Time Step} & \textbf{Attention Weight \& Clinical Interpretation} \\
\midrule
1 & Hour 0 & $0.0806$ --- Baseline admission acuity and shock index \\
2 & Hour 1 & $0.0352$ --- Immediate post-admission fluid/pressor response \\
3 & Hour 2 & $0.0199$ --- Early stabilization response window \\
4 & Hour 3 & $0.0165$ --- Acute stabilization plateau \\
5 & Hour 5 & $0.0149$ --- Mid-acute surveillance checks \\
6 & Hour 4 & $0.0147$ --- Trajectory confirmation \\
7 & Hour 6 & $0.0144$ --- Shift handover and laboratory re-check \\
8 & Hour 8 & $0.0138$ --- Intermediate resuscitation evaluation \\
9 & Hour 7 & $0.0137$ --- Morning clinical review \\
10 & Hour 21 & $0.0136$ --- Day-1 nocturnal stability check \\
\bottomrule
\end{tabular}
}
\end{minipage}
\vspace{1ex}
\subsection{Static Feature Ablation Ranking}
Table~\ref{tab:static_ablation} displays the leave-one-feature-out ablation impact across all 17 static operational and demographic covariates.

\noindent\begin{minipage}{\columnwidth}
\centering
\captionof{table}{Ranking of static features by ablation impact ($|\Delta \text{prob}|$).}
\label{tab:static_ablation}
\scriptsize
\renewcommand{\arraystretch}{0.88}
\setlength{\tabcolsep}{2.5pt}
\resizebox{\columnwidth}{!}{
\begin{tabular}{rlrrp{4.2cm}}
\toprule
\textbf{Rk} & \textbf{Feature} & \textbf{$|\Delta p|$} & \textbf{Share} & \textbf{Clinical Context} \\
\midrule
1 & \texttt{pre\_icu\_transfers} & $0.0926$ & $28.2\%$ & Number of inter-ward handoffs \\
2 & \texttt{primary\_icd} & $0.0542$ & $16.5\%$ & Primary admission diagnosis code \\
3 & \texttt{icu\_speedup\_los} & $0.0315$ & $9.6\%$ & Departmental turnover pressure \\
4 & \texttt{drg\_code} & $0.0302$ & $9.2\%$ & DRG casemix category \\
5 & \texttt{surg\_count} & $0.0226$ & $6.9\%$ & Pre-ICU surgical procedures \\
6 & \texttt{ADMISSION\_LOC} & $0.0215$ & $6.5\%$ & Transfer source (e.g., ED, Clinic) \\
7 & \texttt{FIRST\_CAREUNIT} & $0.0181$ & $5.5\%$ & Specialty ICU unit designation \\
8 & \texttt{age} & $0.0134$ & $4.1\%$ & Patient chronological age \\
9 & \texttt{INSURANCE} & $0.0119$ & $3.6\%$ & Primary healthcare payer class \\
10 & \texttt{ETHNICITY} & $0.0094$ & $2.9\%$ & Demographic profile \\
11 & \texttt{log\_icu\_los} & $0.0075$ & $2.3\%$ & Total log ICU stay duration \\
12 & \texttt{GENDER} & $0.0044$ & $1.3\%$ & Biological sex \\
13 & \texttt{MARITAL\_STATUS} & $0.0039$ & $1.2\%$ & Social support indicator \\
14 & \texttt{ADMISSION\_TYPE} & $0.0025$ & $0.8\%$ & Emergency vs.\ Elective status \\
15 & \texttt{log\_ed\_wait} & $0.0023$ & $0.7\%$ & Pre-ICU ED boarding dwell time \\
16 & \texttt{log\_surg\_gap} & $0.0020$ & $0.6\%$ & Post-operative latency to ICU \\
17 & \texttt{surg\_flag} & $0.0008$ & $0.2\%$ & Binary surgical intervention \\
\bottomrule
\end{tabular}
}
\end{minipage}

\end{document}